\documentclass{article}

\PassOptionsToPackage{numbers, compress}{natbib}

 \usepackage[main, final]{neurips_2025}

\usepackage[utf8]{inputenc} 
\usepackage[T1]{fontenc}    
\usepackage{hyperref}       
\usepackage{url}            
\usepackage{booktabs}       
\usepackage{amsfonts}       
\usepackage{nicefrac}       
\usepackage{microtype}      
\usepackage{xcolor}         
\usepackage[pdftex]{graphicx}
\usepackage{amsmath}
\usepackage{listings}

\title{LLM Agents for Knowledge Discovery in Atomic Layer Processing}
\author{%
  Andreas Werbrouck \thanks{current email: andreas.werbrouck@ugent.be} \\
  \And
  Marshall B. Lindsay\\
  \And
  Matthew Maschmann
  \And
  Matthias J. Young \\
  University of Missouri \\
  Columbia, MO 65201 \\\texttt{matthias.young@missouri.edu}
}

\begin{document}

\maketitle

\begin{abstract}
Independently reasoning large language model (LLM) agents have immense potential for knowledge discovery in materials science. However, it is challenging to discern latent knowledge and summarized knowledge picked up during training from newly discovered knowledge. To work around this issue, we repurpose LangGraph's \verb|tool| functionality to supply agents with a black box function to interrogate without any other objective than correctly describing the system. We provide proof of concept for this approach through a children's parlor game. We then apply the same strategy to show that LLM agents can explore, discover, and exploit diverse chemical interactions in an advanced Atomic Layer Processing reactor simulation using intentionally limited probe capabilities without explicit instructions. For both examples, we discuss path-dependence of results, the importance of persistence and extra context on discovering rare effects.
\end{abstract}

\section{Introduction}

Large Language Models (LLMs) and agentic systems have shown great potential for disruption in many professional fields. In materials science, LLM or agent use focuses either on synthesizing, integrating, and validating domain knowledge \cite{Yanguas2025benchmarking,m2024augmenting, zhang2024honeycombflexiblellmbasedagent,pagel2024validationscientificliteraturechemputation}, or leverages them as components in self-driving labs for materials design, synthesis, or characterization (either computational or real). \cite{jia2024llmatdesignautonomousmaterialsdiscovery,Boiko2023, Tom2024, song2025multiagent}. In most cases, such agents leverage knowledge absorbed during the training phase, are given access to tools and databases, and generally are part of complex, multi-agent workflows designed to reach well-defined objectives (e.g., property optimization). This makes it challenging to separate their ability for knowledge synthesis, discover of latent knowledge, and discovery of completely new knowledge \cite{Wang2023}.



In this work, our goal is to probe the capabilities of LLM agents to interrogate a system and discover completely new knowledge\cite{Stanley2015}. Leveraging tool capabilities, we give the LLM agent a specific system to work with and ask it to provide statements about the rules governing this system \textit{without defining any other objective}. Since the rules are created specifically for this work, minimal background knowledge is required. This system confines the problem in a different way than a question does: we provide confined input and output spaces, leaving room for the agent to experiment in order to discover the rules. 

We demonstrate this through two examples. First, we play a parlor game with the agent: it needs to describe the rules of an alien market through trial and error (the aliens do not sell anything with the letters p or m in the word). Secondly, we study a materials science problem with real-world relevance: the semiconductor industry requires chemical processes able to precisely deposit and/or remove of material, often through surface-mediated chemical reactions  (see appendix \ref{appendix_alp} for background). Discovering and characterizing novel reactions is a time-consuming task. We give the agent access to a complex, simulated chemical reactor with custom hypothetical precursors and reactions, with the task to describe this system. Dedicated in situ pressure and mass change sensors provide the agent with limited, but sufficient probes into the system state, much like a human scientist would not have access to all reactor variables. 

\begin{figure}[ht]
  \centering
  \includegraphics[width=0.6\linewidth]{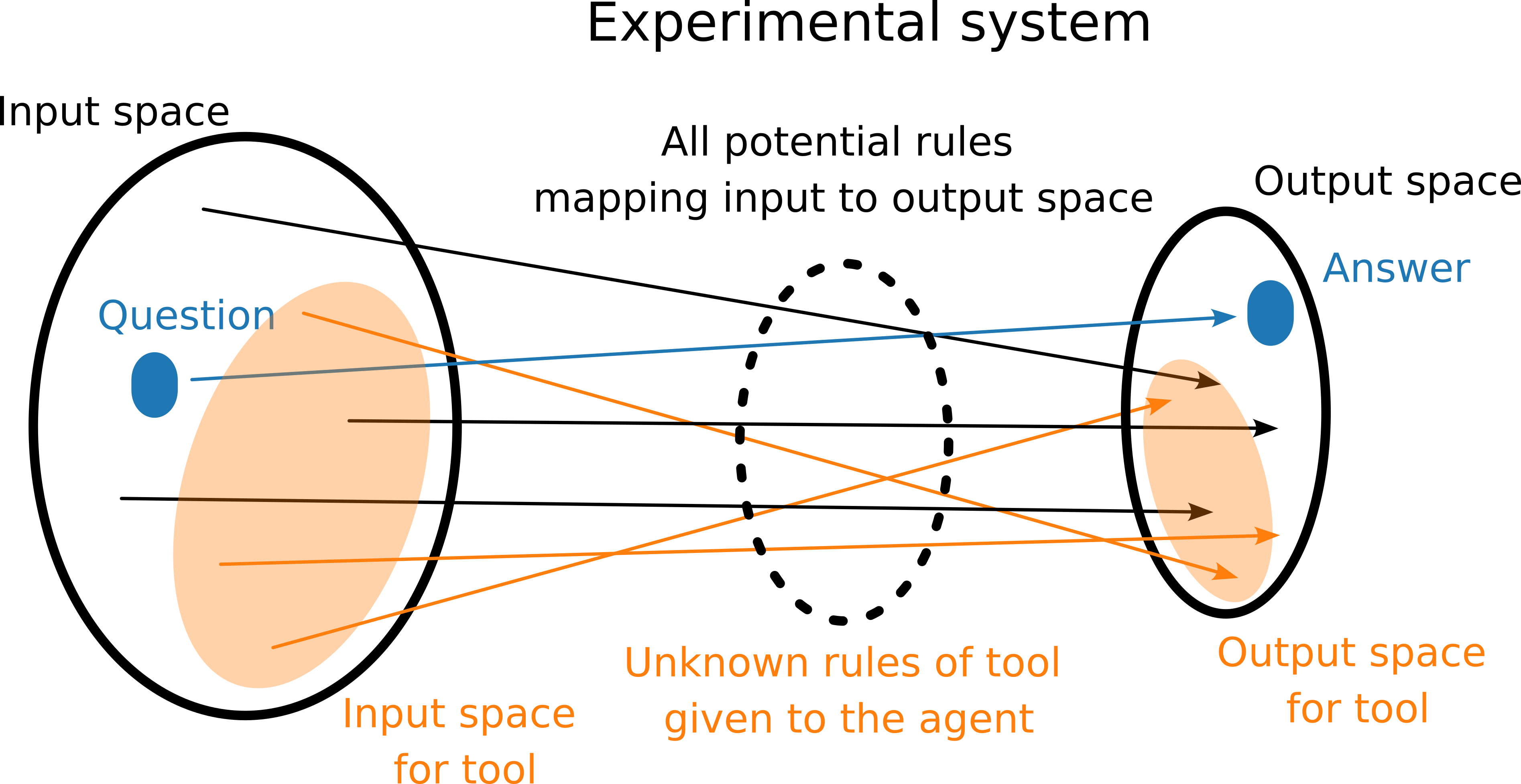}
  \caption{\label{fig0} Discovering rules mapping the input space to the output space (orange) is different from asking an agent to answer specific questions (blue). 
  }
\end{figure}

\section{Methods}


Specific details about agent implementation, prompts and simulation framework can be found in Appendix \ref{appendix_extendedmethods} and \ref{appendix_systemprompts}. The reactor simulation package will be made publicly available at \url{https://github.com/awwerbro/ALDReactor}.


\section{Results}

\begin{figure}
  \centering
  \includegraphics[width=1.0\linewidth]{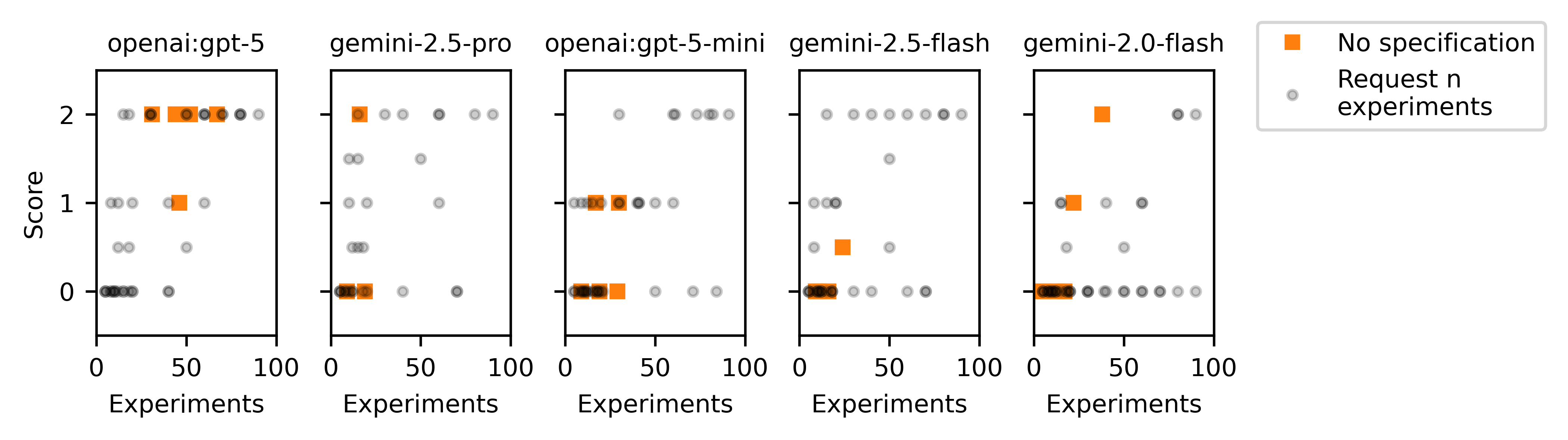}
  \caption{\label{fig2} Initially, most models except for gpt-5 perform poorly at discovering the rules of the Alien market (orange squares). The performance markedly improves upon requesting the agent to perform a defined number of experiments (dark markers).}
\end{figure}

\subsection{Alien market}
We conducted two different experiments with the agent on the Alien market (Figure \ref{fig1}a). We compare flagship models from openAI (gpt-5) and Google (gemini-2.5-pro) with medium-sized models (gpt-5-mini and gemini-2.5-flash) and one older medium-sized model (gemini-2.0-flash). 
In the first experiment (orange squares in Figure \ref{fig2}), we provide the agent with the background story and instruct it to find the rules of the alien market (Scoring is explained in Appendix \ref{appendix_extendedmethods}, Prompt in Appendix \ref{appendix_systemprompts}).

Upon examination of the results, gpt-5 clearly outperforms the other models. However, it does so by conducting significantly more experiments. Other models make some observations, then provide rules that match these observations. This is not indicative of the reasoning capabilities of gpt-5; the other models simply stopped their investigation too soon. In the second set of experiments (Figure \ref{fig2}, dark dots) we explicitly asked the agent to do n experiments, which significantly improved results. We note that pushing models to extend their investigation is a viable strategy to achieve meaningful and generalizable results.

\subsection{ALP reactions}
In this second, more relevant section, we provide the LLM with another system to control and characterize: an advanced ALP reactor simulation. The simulation allows to specify different, fictitious reactions between chemicals beforehand. After specifying the available chemistries, we let the agent control the system, analyze outputs and try to condense the rules of the chemistry using a requested amount of time. 


In the first example, we want the agent to explore what is possible with two fictitious chemicals A and B. Both chemicals react in a self-limiting way, leading to deposition. 
\begin{align}
\text{sA} + \text{B} &\rightarrow \text{sB} + \text{S} + \text{E} \label{eq:reaction1} \\
\text{sB} + \text{A} &\rightarrow \text{sA} + \text{E} \label{eq:reaction2} 
\end{align}

with S a solid, deposited layer, and E a gaseous reaction product. We explore this set of reactions in two experimental configurations (I and II).


In the first experimental configuration (I), transport and reaction kinetics of reaction \ref{eq:reaction1} and \ref{eq:reaction2} were chosen to be very favorable, with 0.5s of exposure to be sufficient to saturate the surface at the starting temperature. An example can be seen in Figure \ref{fig5}. The input space is rather small in this case (only 2 chemicals). As such, in each of 3 iterations, the agent discovered the self-limiting nature of the surface reactions, and exploited these reactions to perform ALD-like depositions. The allocated time of 3600s of experimental time was ample to discover this effect, so with the remainder of their time the agents further explored the space: pulse times and reactor temperature were varied, discovering kinetic limits at low temperature and growth through decomposition.

In the second experimental configuration (II), reaction \ref{eq:reaction2} was made less accessible: reaction kinetics were slower (the prefactor of the Arrhenius equation was 4 times smaller), and the vapor pressure of chemical B was lowered (the B parameter in the Antoine equation was halved). An expert description of the path to fully explore what is possible with the system as a) increasing the temperature of precursor B from 300 K to 550 K, and b) pulsing precursor B for at least 40s. This is an experimentally realistic scenario.  

Three different system prompts were tested with configuration II. The first prompt (IIa) was identical to the prompt used in configuration I, and the agent had as well 3600 s to characterize the system. In every iteration, the agent fully characterized in a low-exposure behavior, characterizing a 'CVD-like' process with very low growth. In the second prompt (IIb) we increased the time available to the agent to 7200s, yet all agents remained stuck in the same local minimum. In the final prompt, we provided a reference value for the QCM measurements ("As a rule of thumb, depending on stoichiometry, a monolayer of material weighs between 10 and 50 ng/cm2""), again with 7200s of experimental time. This time, results were more varied. The first iteration the agent stayed in the low-growth, CVD-like local minimum. The second iteration, it discovered an "ALD-like growth" when the reactor temperature was raised above the decomposition temperature of chemical B. While experimentally correct, it is factually wrong to label this growth mode as ALD. Finally, in the third iteration the agent discovered the `exceptionally slow surface reaction kinetics' \textit{(sic)} and achieved full saturation (completion) of the reactions resulting in higher, controlled growth.



%

Having investigated the a configuration with 2 precursors, we expanded the input space with 2 more chemicals (configuration III). These are summarized in Table \ref{tab_reactions}), departing from reactions \ref{eq:reaction1}-\ref{eq:reaction2}. With these chemicals, more combinations are possible: self-limited deposition (ALD, combining chemical B and C, much like configuration I), self-limited etching (ALE, combining chemical A and C), and passivation through chemical D (ALE, reaction B-C does no longer work, with a pulse of A or an annealing step able to remove the passivating layer). In addition, continuous growth can be achieved by exposing the surface to precursor B and C at the same time (CVD co-dose), and by exploiting the decomposition of precursor C (CVD decomposition, continuous exposure with the reactor temperature above the decomposition temperature of precursor C). Due to the larger experimental space, the kinetics were chosen to be easily discovered.

\begin{figure}
  \centering
  \includegraphics[width=1.0\linewidth]{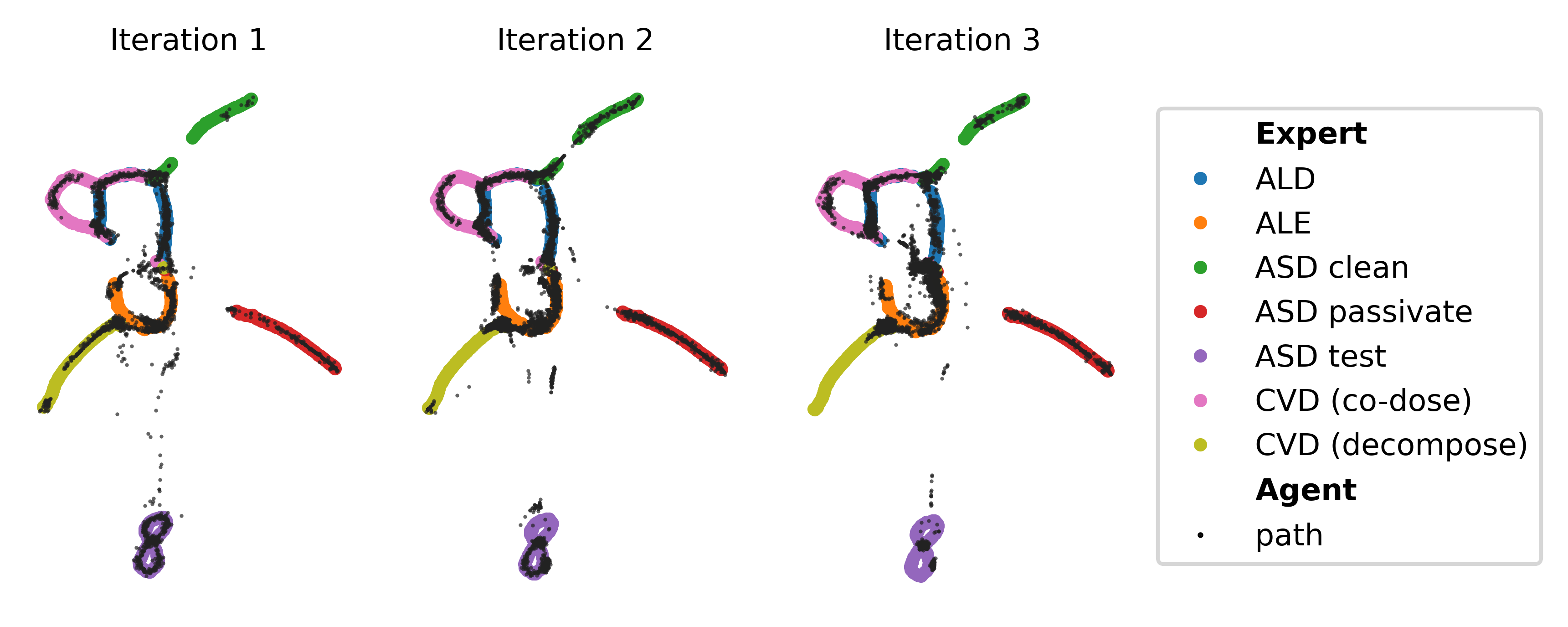}
  \caption{\label{fig6} Visualization of the experimental space visited by the agent in experimental configuration III, as compared to manually labeled experiments conducted by a human with knowledge of the configuration (UMAP projection of the full reactor state in the course of the experiments). In different iterations, different parts of the space are visited.}
\end{figure}

We allowed 7200s of experimental time and employ nudging strategies (see Appendix \ref{appendix_extendedmethods}) to prevent the agent from giving up too soon. In figure \ref{fig6}, we visualize the trajectory of the agent in experimental space. During the simulation, the full reactor state yields a 560-dimensional vector for every timestamp. After subsampling with respect to time to ensure sufficient variation, these vectors are projected into two dimensions using Uniform Manifold approximation and projection (UMAP), which was trained on a labeled reference experimental run, conducted by an expert. If the experimental trajectory covers the entire space indicated by a color, it has brought the reactor in a similar state compared to that of the expert label. The following paragraphs are based on manual investigation of agent statements, the visualization in Figure \ref{fig6} and manual inspection of the QCM/pressure traces. 

In the first iteration the agent discovered ALD with precursors B and C, passivation against B-C with chemical D, and also a CVD process at high temperature. However, it did not recognize/describe ALE, but instead used chemical A as a `cleaning agent' to obtain reproducible results. The temperature stability of the deposited film was also tested. The second iteration discovered ALD, ALE, and used D to passivate the surface against single pulses of B and C, but not exploring passivation against the full B-C cycle. The agent also removed D by annealing the surface (as opposed to pulsing A). Some time was spent pursuing an A-C-B-C cycle, which yields no growth. Co-dosing was not explored. The third iteration focused on etching with A and C. It pursued several dead ends: complex A-C-D cycles with little growth, and the onset of C decomposition in combination with etching through A. It did not describe cyclic reactions of B with C. 


\section{Discussion}

In this section we discuss several common aspects with respect to knowledge discovery between the alien market and the reactor simulation. These apply to agent-based discovery, but are easily generalized to human discovery. We address whether the performed experiments can be described as `knowledge discovery', and conclude with a brief discussion about implications of this work for real-world systems.

First of all, we find that some `rules' or effects are intrinsically harder to find than others, and require a certain amount of (imposed) persistence. Furthermore, some signal has to be detected by the agent to spark `curiosity' and further inquiry. In the absence of signal, two strategies can work: first, one can provide more time/experimental resources to the agent in order to increase the probability of detecting some signal (and nudge the agent to use these resources if necessary). A second strategy can be providing more context in the prompt - which we deliberately limited in our experiments. 

Specifically, for the alien market: based on a list of 10,000 English words \cite{price}, the probability that an item will be rejected by the market (word contains m or p) is 0.3248. On the other hand, if the rejection rule had been based only on the occurrence of 'z', the probability of a 'False' answer would be only 0.0129, leading to longer times before discovery. Finally, imagine the case where the rule is that the aliens only sell objects containing a letter \c c. An agent would have a hard time finding this rule, except if the information that the aliens prefer to communicate in French was available from the beginning. 
For the chemical reactions, ALP configurations I and II illustrate the effect of the availability of context, combined with more experimental time to get out of local minima: identification of a self-limiting ALD reaction only happened after providing the agent with reasonable QCM ranges in configuration IIc. 

Another general observation is that discovery of the rules is highly path-dependent. In the alien market, it was extremely common for any agent to start with "apple". In some cases, this led to construction \textit{and} validation of the rule that "any word with double p is rejected". Increasing model temperature could lead to more diversity in the generation of starting points and subsequent paths. For the ALP reactor, configuration III also exhibits path-dependence. In the 3 iterations, the agents explored different parts of the system. This can inspire swarm-like strategies, in which a supervising agent summarizes results from different sub-agents exploring a similar system. We intentionally randomized the order of the chemicals to avoid path-dependent bias. Tweaking model temperature could also be a strategy to increas path-induced outcome variety. 

Since both the Alien market rules and the chemical reactions to be discovered were hypothetical and conjured up by us, we consider this a proper example of knowledge discovery: the agent had no prior knowledge about the used chemicals to rely on, except what we provided. More discussion on the alien market experiment can be found in Appendix \ref{appendix_alien}. More discussion on avoiding priming the agent through the prompts can be found in Appendix \ref{appendix_alp_discussion}. 

In these two examples we provided the LLM with specific effects to discover. However, in reality, the number and complexity of input-output combinations (and hence, the opportunities for failure \textit{and} success) are seemingly endless. 
As such, we want to stress that success is defined by the way the results are interpreted. Without prior knowledge, it is hard or even impossible to \textit{optimize} an experimental trajectory (the reference recepe in Table \ref{tab_reference_recipe} could be considered optimized). However, even an inefficient path can yield results that turn out to be interesting in a different context. The ability to conduct experiments without predefined objective, besides exploring a system, could enable construction of comprehensive scientific databases without the bias toward success present in the scientific literature. 

This leads to an interesting tradeoff: in an experimental system with real, well-described chemicals, agents could (but not necessarily should) be augmented with more knowledge, strategies and/or tools. However, \textit{not} knowing something, and ignoring or critical examination of prior knowledge could lead to more novel paths. A similar tradeoff exists with respect to experimental constraints: these should be in place for safe use of equipment, yet sufficiently broad and flexible to allow the agent to explore beyond well-trodden paths.

Finally, we note that related work has been carried out by Duan \textit{et al.}. In that work, LLM's were allowed to perturb simulated biological systems and tasked with modelling the reactions in python \cite{duan2025}. 

\section{Conclusions}
In this work, we use two simulated experimental systems, of which one has high relevance for the field of materials science, to show that LLM agents can a) interrogate unknown systems and reason moderately well on the outcomes, when prompted to use sufficient experimental resources b) pursue complex and interesting ideas as a result of initial observations c) summarize their findings as general statements about the system under study, even without other specific objectives.


As a materials science community, we need to broaden the scope of our efforts with respect to using AI for our research. Tremendous progress has been made to ensure factual and correct LLM responses, to discover latent knowledge in literature, and to optimize for specific objectives. However, independent discovery in data-poor conditions for yet-to-be-discovered effects is an exciting area where AI/ML can also be of use. 


\newpage
\bibliographystyle{unsrtnat}
\bibliography{references}

\newpage
\appendix

\section{Atomic Layer Processing}
\label{appendix_alp}
\renewcommand{\thefigure}{A\arabic{figure}}
\renewcommand{\thetable}{A\arabic{table}}
\setcounter{figure}{0}
\setcounter{table}{0}
Atomic Layer Processing (ALP\cite{ashurbekova2025atomic}) encompasses several techniques at the forefront of the semiconductor industry including atomic layer deposition (ALD) \cite{puurunen2005surface}, atomic layer etching (ALE)\cite{fischer2021thermal}, and area selective deposition (ASD)\cite{parsons2020area}. These techniques harness self-limiting gas-surface chemistries to enable atomic-scale control over thin-film stacks. In ALD, surfaces in the reactor are exposed to multiple cycles of sequential pulses and purges of two chemicals (e.g., trimethylaluminum and water). Due to the self-limiting nature of the interactions, complex, three-dimensional substrates can be coated in a conformal and uniform way \cite{Cremers2019}. This ability has been a key component of the semiconductor manufacturing toolkit for decades, and also finds application in other industries. ALE behaves practically the same, with the only difference that the material is \textit{removed} in a self-limiting way instead of deposited. ASD, finally, employs the chemical passivation of certain surfaces to selectively control chemical access (e.g., of ALD) on specific surfaces, while protecting others. However, discovering, characterizing, and verifying novel chemistries and precursor-surface interactions is a time-consuming and expensive effort. Therefore, it is an excellent testbed for ML/AI approaches to speed up discovery. Earlier work has focused on (among others) predicting \cite{YanguasGil2022} and optimizing saturation times (see \cite{Paulson2021} and references therein). 

\section{Extended Methods}
\renewcommand{\thefigure}{B\arabic{figure}}
\renewcommand{\thetable}{B\arabic{table}}
\setcounter{figure}{0}
\setcounter{table}{0}

\begin{figure}
  \centering
  \includegraphics[width=0.8\linewidth]{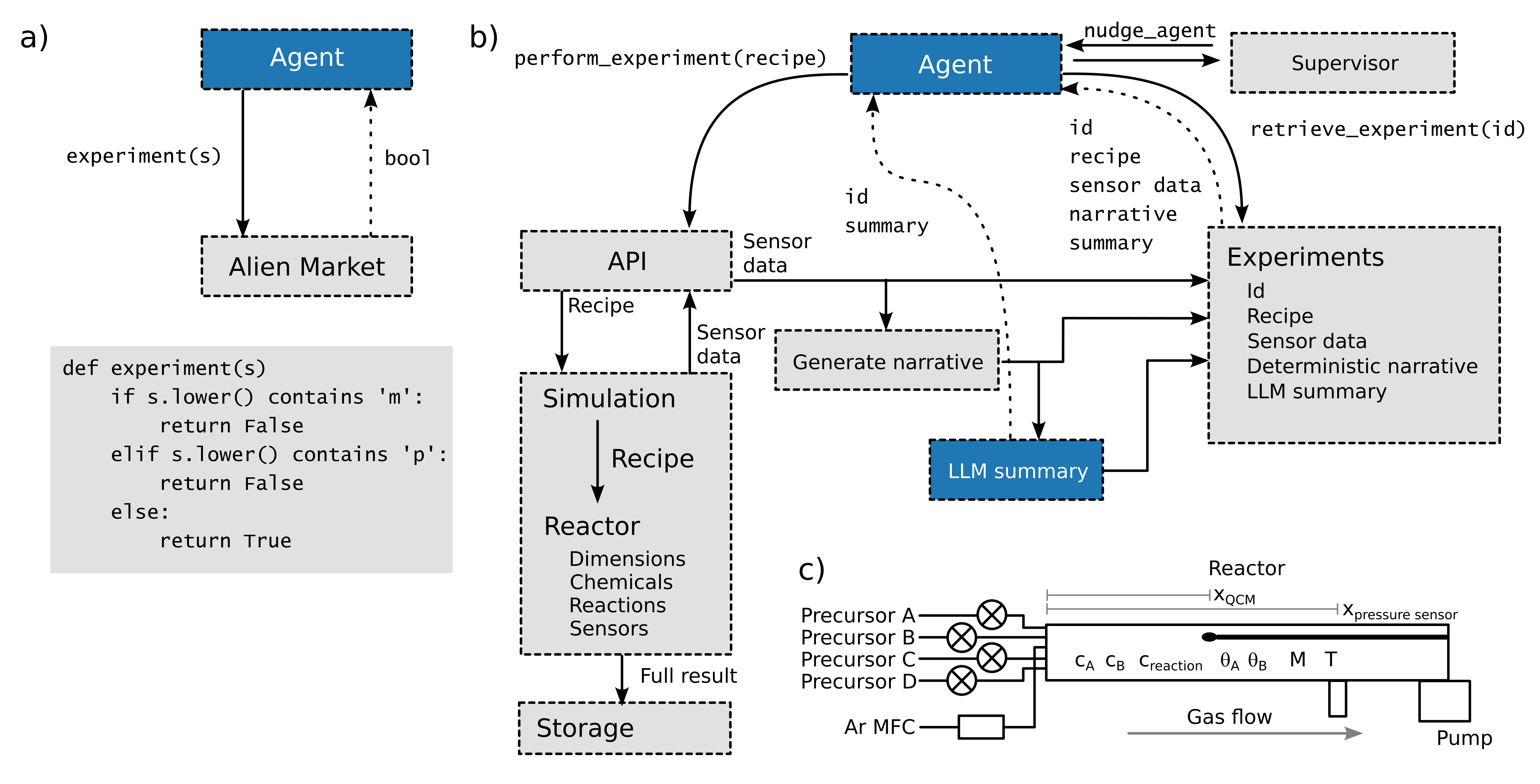}
  \caption{\label{fig1} a) graph for the first example (Alien market) b) graph for the second example (ALP reactor) LLM-based components are shown in blue, while deterministic components are grey. c) schematic reactor layout.}
\end{figure}

\label{appendix_extendedmethods}

\subsection{Alien market: LangChain implementation}
A LangChain ReACT agent \cite{yao2023react} was used in the first example (Alien market) with different models as indicated. The system prompt was identical for all models and can be found in Appendix \ref{appendix_systemprompts}. The agent is given only one tool: the function to interrogate (Figure \ref{fig1}a). 


\subsection{Alien market: experimental conditions}
In the first set of experiments (Figure \ref{fig2}, orange dots), every agent was tested five times to figure out the rules without specifying anything with regards to the number of experiments. In the second set of experiments (Figure \ref{fig2}, dark dots) we explicitly asked the agent to do n experiments (n = 5, 8, 10, 12, 15, 18, 20, 30, 40, 50, 60, 70, 80, 90 experiments, with three iterations per condition). The total number of experiments was returned with each result. 

\subsection{Alien market: evaluation}
The solutions were evaluated through another LLM, but after some unreliable scores we graded the conclusions by hand. The ground truth rules are that if the string contains m or p (case insensitive), it is impossible to buy it on the market. If the LLM gave a completely wrong answer, no points were awarded. One point was awarded for correct identification of a letter. Another point was awarded for the correct identification of the other letter. In some cases the LLM added a third rule about another letter being disallowed. For these unnecessary extra rules we subtracted 0.5 points. This led to 1.5 points for a statement like "words containing the letters p, m or l are disallowed" and a score of 0.5 for a statement like "words containing the letters p/P or l/L are disallowed". 

\subsection{ALP reactor: LangChain implementation}
In the second example (Atomic Layer Processing), a similar ReAct Agent was used (model: gemini 2.5 pro). We considered more complex multi-agent architecture but decided against it since the objective of this study is probing the discovery capabilities of existing LLMs, not to innovate networks of LLMs to improve this function. However, we did make some decisions to filter the information reaching the Agent (Figure \ref{fig1}b). More specifically, the agent has two tools now. The \verb|perform_experiment| tool accepts a recipe string (a tab-based table, see Appendix \ref{appendix_systemprompts} and \ref{appendix_equations}) which is passed on to an API on top of the reactor simulation. The string format is checked and the simulation is carried out. The reason for this format is that the physical reactors in our lab also accept this format, so it would be trivial to add an API layer onto the physical reactor control and swap the endpoints to conduct physical experiments controlled by an LLM. Once the experiment finished, the sensor (pressure and mass at specific locations), valve and temperature data is returned with 0.1s resolution. A deterministic narrative (string) is constructed from this data, which is then summarized by a second LLM (gemini 2.5 flash) into a short, high-level description of what happened in the reactor during the process. The entire experiment (id, sensor data, narrative, summary) is stored for later use, but is not dumped on the agent. However, if desired, the second tool, \verb|retrieve_experiment| passes the complete information to the main Agent. The complete experiment, including discretized pressure and coverage profiles across the reactor, is saved for reference but is not made available to the agent.

In the system prompt (Appendix \ref{appendix_systemprompts}) a configuration-dependent description of the reactor and available chemicals is provided, with instructions on how to provide strings to control this system (see Appendix \ref{appendix_equations} and \ref{appendix_systemprompts}). Here, we exploit the capability of LLMs for single-shot learning with respect to tool use \cite{brown2020}. Gas flow rate, temperatures and valves can be changed independently. The output presented to the agent consists of a pressure signal and a Quartz Crystal Microbalance (QCM) signal versus time during the experiment, similar to what a human experimenter would see. By design, neither of the output signals allows the agent to unambiguously distinguish between the chemicals.

We also provide the agent with a defined amount of reactor time to use, to ensure continued experimentation, much like the Alien Market example. Initial experiments indicated that agents consistently used significantly less than the allocated/requested experimental time in all cases. Different prompting strategies were tested, but in contrast to the earlier example, we could not force the agent to use all its time. Eventually, a simple system kept the agent in the loop, prompting it to continue research until 90\% of the available time was used. 


\subsection{ALP reactor: simulation and system of equations}
\label{appendix_equations}
\subsubsection{Reactor, precursors and sensors}
The reactor is modeled as a tube with diameter $d$ and length $L$, discretized into $N$ sections kept at temperature $T$. The reactor is considered to be heated uniformly, so this temperature is used to evaluate all equations. Temperature changes happen immediately.

A mass flow controller (MFC) at the inlet defines carrier gas flow $f$ (volume per unit time) entering the reactor. At the outlet, a pump is defined having a nominal pumping speed S (m$^3$/s), and base pressure (Pa) the pump can attain without load. However, at base pressure, the pump displacement will be much lower than the nominal speed and net to zero: indeed, the nominal pumping speed depends on the pump type, pump health, and pressure. We approximate the pump speed as a linear function of pressure below a threshold value $p_{\text{thresh}}$ (which is pump-dependent) such that

\begin{equation}
S_\text{pump}(p) = \begin{cases}
S_{\text{nom}} & \text{if } p > p_{\text{thresh}} \\
S_{\text{nom}} \cdot \frac{p - p_{\text{min}}}{p_{\text{thresh}} - p_{\text{min}}} & \text{if } p_{\text{min}} \leq p \leq p_{\text{thresh}}
\end{cases}
\end{equation}

For the MFC the displaced volume per second also depends on the pressure, through

\begin{equation}
    S_\text{MFC}(p) = f_{m^3/s} \frac{ T_{reactor} \cdot 101325 \text{ Pa}}{p \cdot 273.15 \text{ K}} 
\end{equation}

In equilibrium, by definition, we have $S_\text{pump}(p) = S_\text{MFC}(p)$, such that in the transition region the pressure at the pump inlet $p_\text{inlet}$ as a result of a specific flow is given by finding the root of a second-order polynomial.

Additionally, for a gas with dynamic viscosity $\mu$ the pressure drop can be calculated through the Hagen-Poiseuille equation

\begin{equation}
    \frac{dp}{dx} = -\frac{8 \pi \mu f}{A^2}
\end{equation}

However, this pressure drop is minimal under operating conditions. The local (carrier) gas velocity can then be calculated from the local displaced volume per time unit (at reactor conditions) through dividing by the reactor cross-sectional area. The velocity and pressure drop are interdependent, but for all practical purposes they will be considered constant over the length of the reactor. We approximate that the gas velocity is only influenced by the inert gas flow, and not by any other chemicals.

\textit{Gas-phase molecules} are defined through their total molar mass $m_i$, and a diffusion coefficient $D$. Diffusion of precursors in laminar flow (low vacuum and/or macroscopic features) can be calculated through Chapman-Enskog theory \cite{MASON1970155}, while in the molecular flow (high vacuum and/or microscopic features) Knudsen theory can be used \cite{Knudsen1909}. In the transient region between laminar and molecular flow, the effective diffusion coefficient $D_i$ (m$^2$/s) for each precursor can be calculated through the Bosanquet relation \cite{Poodt2017, Ylilammi2018}. Here, we are firmly in the laminar flow regime (1 Torr, macroscopic reactor).
Additionally, \textit{Precursors} have a decomposition temperature and parameters for the Antoine equation to determine their vapor pressure when in the bubbler. Each precursor is contained in a bubbler to separate extrinsic, environment-dependent properties (e.g., vapor pressure, temperature) from intrinsic properties (e.g., Antoine coefficients). For every precursor, a discretized pressure profile is defined across the reactor. Solving this pressure profile is one of the major tasks of the simulation.

\textit{Surfaces} are also defined in a discretized, dimensionless way across the reactor (surface coverage $0 < \theta < 1$). They have a molecular weight and a surface density of sites, defined as $\sigma$ (mol/m$^2$).

A reaction is then defined between precursors and surfaces with reaction rate $k_i$, such that:
\begin{equation}
    r_i = k_i  c_i  \theta_i  \sigma
\end{equation}

Once $r_i$ is known, the participating species can be updated. For example the reaction between molecule C (e.g., trimethylaluminum) and surface sB (OH) will have a specific reaction rate for every position in the reactor. This reaction results in the solid S being created with rate $r$. It is also possible to define stoichiometries with a reaction. The reaction rate $k_i$ can be modeled as an arbitrary, temperature-dependent equation. In this case first-order Arrhenius kinetics were used.A

Once the set of reactions has been modeled, the reactive transport of gaseous species $i$ (precursors and reaction products) can then be modeled with

\begin{equation}
    \frac{\partial c_i (x, t)}{\partial t} = D_i \frac{\partial^2c_i}{\partial x^2} - v \frac{\partial c_i}{\partial x} - \frac{4}{d}\sum_j r_{ij} 
    \label{eq_c}
\end{equation}

and the change in coverage is 
\begin{equation}
    \frac{\partial \theta_i}{\partial t} = \frac{1}{\sigma_i}\sum_j r_{ij}
    \label{eq_th}
\end{equation}

where $r_{ij}$ indexes over all $j$ reactions involving species $i$ and the sign of each $r$ indicates consumption or creation.

This flexible system allows for arbitrary functional forms of $k$ (e.g., $k(T) = Ae^{E/k_BT}$, or 
\begin{equation}
    k(T) = \begin{cases} 0 & T < 500K \\100 T& T \geq 500K\end{cases}
\end{equation}

This includes irreversible processes and reversible physisorption processes. Since the latter cause significant overhead, they were not included in the simulations. For easy setup, a reactor description including reactor dimensions, bubblers, precursors, reaction products, surfaces, solids and sensors can be specified as a .json file and loaded.

During a simulation, the coupled system of differential equations \ref{eq_th} and \ref{eq_c} is solved together with other relevant parameters (precursor consumption in the bubbler, etc.). Equations \ref{eq_th} and \ref{eq_c} are similar to those reported in earlier work simulating reaction-diffusion into high aspect ratio features in the context of conformality research \cite{YanguasGil2012, Ylilammi2018, Cremers2019, Gonsalves2024}. Key differences between the reaction-diffusion system considered in conformality research is the addition of the advection term (the effect of inert carrier gas velocity) and a change in outlet boundary conditions because of the pump. In our simplified view, stoichiometry and chemical specifics of the reaction are not taken into account. 

While the system of equations describes reactive transport, the boundary conditions determine how the actual system evolves. At the inlet, we assume $c_i(0) = 0 \text{ or } \frac{p_i}{RT}$ depending on opening or closing of the valves, and calculate the first and second order gradients based on those values (Dirichlet boundary condition). At the end of the reactor, we assume all gas is pumped, such that there is no accumulation of precursor, leading to a natural boundary condition from the equations.

The reactor configuration allows for placement of one or more pressure sensors across the length of the reactor. The total pressure recorded by these sensors is then passed on to the agent.

\textbf{QCM calculations}
We have direct access to the deposited mass per surface area in our calculations. As such, it is convenient to define one or more quartz crystal microbalance (QCM) sensors across the reactor. Before passing the data to the agent, units are converted from g/m$^2$ to ng/cm$^2$.

\subsubsection{Recipe}
\begin{table}[ht]
    \centering
    \begin{tabular}{cccccl}
        \hline
        \textbf{Cycles} & \textbf{Type} & \textbf{\#} & \textbf{Action} & \textbf{Wait (s)} & \\
        \hline
        1 & M & 1 & 50 & 0 & \textit{\# MFC 1 @50 SCCM}\\
          & V & 2 & 0 & 0 & \textit{\# close valve 2}\\
          & V & 3 & 0 & 10 & \textit{\# close valve 3} \\
        5 & V & 2 & 1 & 1 & \textit{\# open valve 2, wait 1s}\\
          & V & 2 & 0 & 10 & \textit{\# close valve 2, wait 10s}\\
          & V & 3 & 1 & 1 & \textit{\# open valve 3, wait 1s}\\
          & V & 3 & 0 & 10 & \textit{\#close valve 3, wait 10s}\\
        \hline
    \end{tabular}
    \caption{Contents of a typical recipe file pulsing precursor B and C. The first three lines ensure that the reactor is in a suitable state: the Ar flow is switched to the desired setting (line 1), valve 2 is closed and valve 3 is closed. After the third line, a 10s wait time is enacted. Then five cycles of the following sequence are delivered: (pulse 1) Valve 2 is opened, 1s wait, (purge 1) valve 2 is closed, 10s wait (pulse 2) valve 3 is opened, 1s wait (purge 2) valve 3 is closed, 10s wait. These last four lines are repeated five times. Multiple cycle conditions can be scheduled after each other in this way. The output of this recipe (with reactor configuration III) is shown in Figure \ref{fig5}}
    \label{tab_recipe}
\end{table}

\begin{figure}
  \centering
  \includegraphics[width=1.0\linewidth]{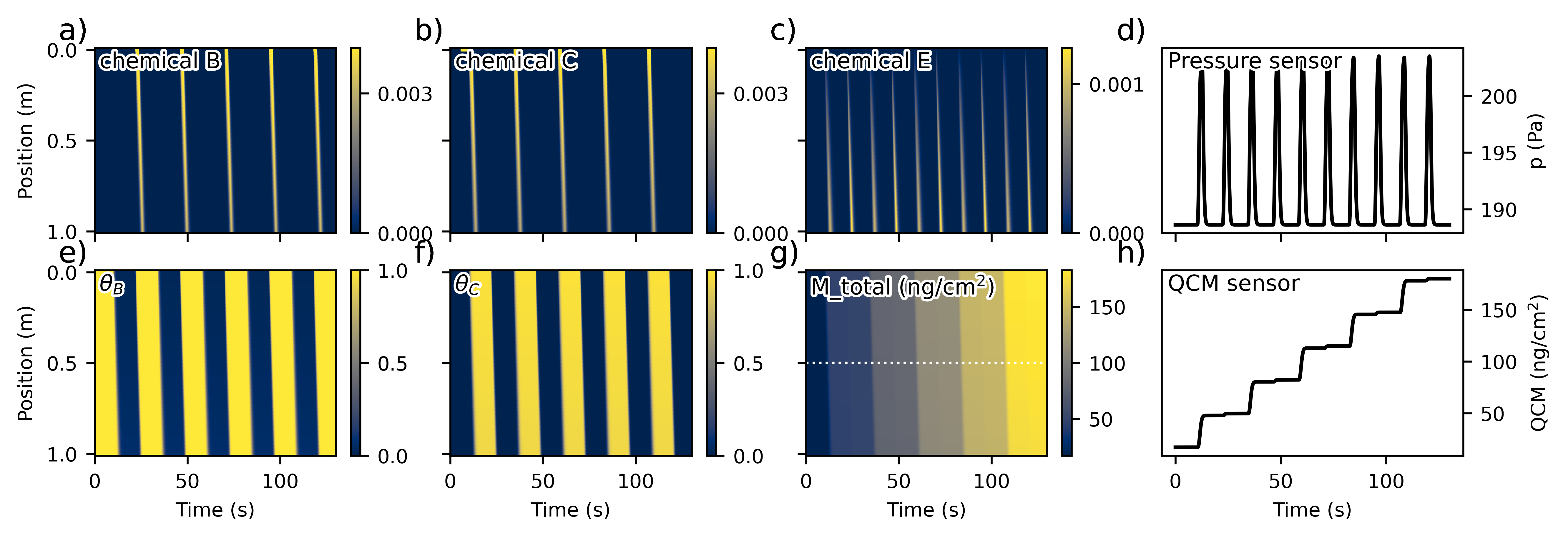}
  \caption{\label{fig5} Full concentration profiles  (a, b, c), surface coverage profiles (e, f), mass deposition profiles (g) generated during an experiment pulsing A and B (B and C in the full ALP example), generating gaseous reaction product E and solid S. The dotted line in g) shows the position of the QCM sensor. Sensor data available to the agent is displayed in d) and h).}
\end{figure}

The home-built reactors in our group are controlled through LabVIEW. They can be operated in various ways: manually; by defining typical ALD sequences (AB, mAnB or mABnCD); or through submitting a tab-separated .txt file containing a table with process steps (an example is given in Table \ref{tab_recipe}). This way of passing process steps to the reactor offers great flexibility and interoperability, as reactor control actions can be scheduled arbitrarily (safety checks are built in). To allow for accurate comparison, we decided to control the simulation through files following the same convention. This way, the simulation becomes an effective practice playground for agents. If confidence in the agents grows, switching from simulation to control of physical reactors will be smooth.

\subsubsection{Simulations}
Once a reactor object with precursors and sensors is initialized, a recipe can be simulated on this object. The recipe allows for natural segmentation of the simulation: each process line is executed as a sub-simulation, starting from the end state of the previous sub-simulation (concentrations, $\theta$, M). This way, it is ensured that all steps are processed correctly and numerical instabilities as boundary conditions change sharply at the reactor inlet are avoided. At the end, the segments are concatenated.

On a larger scale, the reactor object retains the state of $\theta$, M and QCM signal between recipe runs - reaction product concentrations are not retained as the reactor would naturally equilibrate between processes.

\subsection{ALP reactor: experimental condititions}

A schematic of the reactor is displayed in Figure \ref{fig1}c. More information and relevant equations for the reactor simulation framework can be found in Appendix \ref{appendix_equations}. An example of a full process input is given in Table \ref{tab_recipe}, where the full output is shown in Figure \ref{fig5}.

The full set of reactions available to the model in the full ALP reaction is given in Table \ref{tab_reactions}.

\begin{table}
\centering
\begin{tabular}{c|cccc}
 & \textbf{sA (AlF$_3$)} & \textbf{sB (-OH)} & \textbf{sC (-CH$_3$)} & \textbf{sD (-CH$_2$CH$_3$)} \\
\hline
\textbf{A (HF)}       
  & -- 
  & $sA + B \;(-S)$ 
  & \textcolor{red}{$sA + E \;(-S)$}
  & $sA + E \;(-S)$ \\
\textbf{B (H$_2$O)}   
  & -- 
  & -- 
  & \textcolor{blue}{$sB + E$} 
  & -- \\
\textbf{C (TMA)}      
  & \textcolor{red}{$sC + F$}
  & \textcolor{blue}{$sC + E + S$} 
  & -- 
  & -- \\
\textbf{D (EtOH)}     
  & -- 
  & $sD + B$ 
  & $sD + E$ 
  & -- \\
\end{tabular}
\newline\newline
\caption{\label{tab_reactions} Self-limiting reactions available to the system. The rows represent the gas-phase chemicals participating in reactions with the surfaces represented in the columns (prefix s). E and F are gas-phase reaction products, and S is the solid material deposited. The actual chemicals (HF, H$_2$O) etc. and surface groups (-OH, -CH$_3$) serve as references to existing processes for the familiar reader, but these are not mentioned to the agent and merely serve as inspiration. Decomposition reactions for C + sX $\,\to\,$ sC + S (linear with T above 600K) and D + sX $\,\to\,$sB + E (linear with T above 500K) are also implemented but not shown in the table.}
\end{table}

\subsection{ALP reactor: UMAP projection and evaluation}
To make the visualization in Figure \ref{fig6}, the simulation recipe in table \ref{tab_reference_recipe} was carried out. 

\begin{figure}
  \centering
  \includegraphics[width=\linewidth]{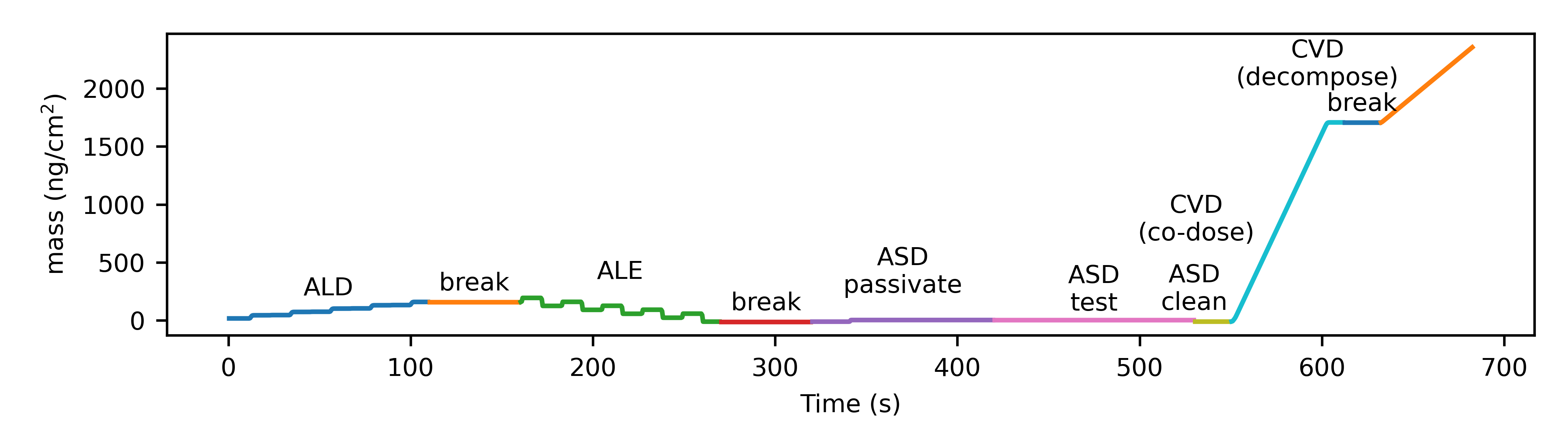}
  \caption{\label{fig7} QCM signal and labels during the reference run.}
\end{figure}

QCM output during this run is displayed in Figure \ref{fig7} The full output of the reactor state during the simulation was a 201371 $\times$ 560 array (not including time). This array was subsampled using k-means clustering with 100 centroids per labeled condition to ensure proper sampling of each condition. These 560-dimensional centroids were then transformed to 2 dimensions using UMAP.

\begin{table}[ht]
    \centering
    \begin{tabular}{cccccl}
        \hline
        \textbf{Cycles} & \textbf{Type} & \textbf{\#} & \textbf{Action} & \textbf{Wait (s)} & \\
        \hline
        1 & T & 0 & 500 & 0 & \textit{\# setup (0 s)}\\
          & T & 1 & 300 & 0 & \\
          & T & 2 & 300 & 0 & \\
          & M & 1 & 50 & 0 & \\
          & V & 1 & 0 & 0 & \\
          & V & 2 & 0 & 0 & \\
          & V & 3 & 0 & 0 & \\
          & V & 4 & 0 & 0 & \\
        5 & V & 2 & 1 & 1 & \textit{\# ALD (110 s)}\\
          & V & 2 & 0 & 10 & \\
          & V & 3 & 1 & 1 & \\
          & V & 3 & 0 & 10 & \\
        1 & V & 1 & 0 & 50 & \textit{\# break (50 s)}\\
        5 & V & 1 & 1 & 1 & \textit{\# ALE (110 s)}\\
          & V & 1 & 0 & 10 & \\
          & V & 3 & 1 & 1 & \\
          & V & 3 & 0 & 10 & \\
        1 & V & 1 & 0 & 50 & \textit{\# break (50 s)}\\
        1 & T & 0 & 400 & 10 & \textit{\# ASD passivation (100 s)}\\
          & T & 4 & 550 & 10 & \\
          & V & 4 & 1 & 60 & \\
          & V & 4 & 0 & 20 & \\
        5 & V & 3 & 1 & 1 & \textit{\# ASD test (110 s)}\\
          & V & 3 & 0 & 10 & \\
          & V & 2 & 1 & 1 & \\
          & V & 2 & 0 & 10 & \\
        1 & T & 0 & 700 & 0 & \textit{\# ASD clean (anneal) (20 s)}\\
          & T & 4 & 300 & 10 & \\
          & T & 0 & 450 & 10 & \\
        1 & V & 3 & 1 & 1 & \textit{\# CVD (co-dose) (52 s)}\\
          & V & 2 & 1 & 50 & \\
          & V & 2 & 0 & 1 & \\
          & V & 3 & 0 & 10 & \textit{\# break (10 s)}\\
        1 & T & 0 & 700 & 20 & \textit{\# break (20 s)}\\
        1 & V & 3 & 1 & 50 & \textit{\# CVD (decomposition) (51 s)}\\
          & V & 3 & 0 & 1 & \\
        \hline
    \end{tabular}
    \caption{Reference recipe used for UMAP projection and evaluation, demonstrating various ALP processes including ALD, ALE, ASD passivation and cleaning, and CVD operations. Comments indicate labels that were added afterwards. The 'break' label was removed from the visualization. }
    \label{tab_reference_recipe}
\end{table}

\section{System prompts}
\label{appendix_systemprompts}
\renewcommand{\thefigure}{C\arabic{figure}}
\renewcommand{\thetable}{C\arabic{table}}
\setcounter{figure}{0}
\setcounter{table}{0}

\subsection{Alien Market}
prompt 1:
\begin{lstlisting}
"""
You are on an extragalactical planet with a market where you can buy virtually everything. Everything? 
No - the aliens have particular rules about what they sell and what they do not sell. Your goal is to figure out what their rules are. 
Whenever you are calling the tool, explain why. 
Make sure you verify your responses.
"""
\end{lstlisting}

prompt 2 (specify number of calls)
\begin{lstlisting}
"""
You are on an extragalactical planet with a market where you can buy virtually everything. Everything? 
No - the aliens have particular rules about what they sell and what they do not sell. Your goal is to figure out what their rules are using the tool. 

Whenever you are calling the tool, explain why. 

Make exactly {n_calls} toolcalls before reporting your final answer. If you don't know, indicate that in the final answer. 
As you are doing experiments, keep refining your hypothesis. Think along the way.
"""
\end{lstlisting}

\subsection{ALP discovery}

\textbf{summarize narrative agent (same in all cases)}
\begin{lstlisting}
"""
You will be given a description of what happens during the experiment.
Return a concise summary, highlighting repeating or unique behavior.
"""
\end{lstlisting}

The system prompt for the main agent in example 2 is constructed from reactor\_prompt, recipe\_prompt and then a prompt that is slightly adapted for the specific run (1,2 or 3).

\textbf{reactor\_prompt}
\begin{lstlisting}
"""
REACTOR SETUP
The reactor is a tube of 1m long and 0.05m diameter with valves at one end, and a pump at the other end. 

- Valve 1: Connected to chemical A (decomposes above 600K)
- Valve 2: Connected to chemical B (decomposes above 600K)
- Valve 3: Connected to chemical C (decomposes above 600K)
- Valve 4: Connected to chemical D (decomposes above 500K)
- MFC 1: Controls Ar gas flow (0-100 sccm)

You have access to a QCM (measuring in ng/cm^2) and a pressure gauge (measuring in Pa) which are both located halfway this tube.

"""
\end{lstlisting}

\textbf{recipe\_prompt}
\begin{lstlisting}
"""
No further information is available about these chemicals, but you can assume their use is safe within the specified limits of the reactor, although they may not behave as desired. 
The safe operating pressure of the reactor is around 200 Pa. Try to keep the pressure below 500 Pa.
The safe temperature for the reactor is about 700K. Stay below this temperature.

To manipulate the reactor, use the following recipe format

RECIPE FORMAT
- Each line has five rows: [cycles] action component_id setting wait_time
- cycles: optional number (if omitted, continues current cycle)
- action: 'M' for MFC, 'V' for valve, 'T' for temperature
- component_id: 1, 2, etc.
- setting: for valves 0/1 (close/open), for MFC 0-1000 (sccm)
- wait_time: seconds to wait

Example experiment string:
"
1   M   1   50  0.      # Set purge gas to 50 sccm, wait 0s. 
    T   0   500 0.      # set reactor temperature (number 0) to 500K
    T   1   350 0.      # set temperature of bubbler 1 (containing chemical A) to 350 K      
    V   1   0   0.      # make sure valve 1 (opening access to bubbler 1) is closed before you start the process, wait 0s.
    V   2   0   0.      # make sure valve 2 is closed before you start the process, wait 0s.
    V   3   0   0.      # make sure valve 3 is closed before you start the process, wait 0s.    
    V   4   0   10.     # make sure valve 4 is closed before you start the process, wait 10s.                                          
5   V   1   1   1.      # open valve 1, wait 1s
    V   1   0   10.     # close valve 1, wait 10s
3   V   2   1   2.      # open valve 2, wait 1s
    V   2   0   10.     # close valve 2, wait 10s
"

The first 7 lines are executed only once (the first line has a 1). They serve to make sure that the reactor is in the right conditions). However, you do not need to do this every time.
After setting these values immediately after each other (wait time 0) the reactor waits 10s (seventh line).
Then the process is ready to perform some pulses: the next two steps open and close a valve (open, wait 1s as the gas flows in the reactor, close, 10s purge). This is repeated five times. 
Finally, the last two lines are repeated 3 times, this time with 2 seconds exposure and 10s wait for chemical B. The other chemicals work in the same way.
Note that the Ar remains flowing at 50 sccm all the time.
You can make the recipe as long as you want.
"""
\end{lstlisting}

\textbf{ALP example}

\begin{lstlisting}
You are research assistant with access to a reactor tool that can investigate gas-surface reactions.
    Your goal is to explore what is possible with this system. 

    OBJECTIVE
    You have {time_available}s of reactor time to perform your experiments.
    Generate a list of statements about the system under study, based on the experiments you did. 
    Do not describe individual experiments, but identify general patterns, as specific as possible (mention values where you can).
    You will be evaluated on the correctness, diversity and generality of these statements, so explore the system thoroughly.

    TOOLS
    conduct_experiment: this allows you to perform experiments by submitting a recipe. You will receive a summary of the experiment.
    retrieve_experiment: you can inspect raw data for an earlier experiment by using this tool with the experiment id.
    {reactor_prompt}
    {recipe_prompt}

    IMPORTANT NOTES
    - it is advisable to close every valve after opening it - but if you have good reasons to do so this is not strictly necessary.
    - Do not use chemicals above their decomposition temperature unless you have a good reason to do so.
    - Surface changes can be persistent in between runs. One pulse a couple experiments ago may still impact your current experiments.
    - Your QCM signal is sensitive enough to detect meaningful changes in the surface. 
    - Your pressure sensor measures chemical A/B/C/D and gaseous reaction products indiscriminately.

    Explain your thinking along the way.
\end{lstlisting}

\section{More observations on the alien market}
\label{appendix_alien}
\renewcommand{\thefigure}{D\arabic{figure}}
\renewcommand{\thetable}{D\arabic{table}}
\setcounter{figure}{0}
\setcounter{table}{0}


Other relevant notes on the alien market are: a) it is sufficiently small that the entire experimental history easily fits in the context window, and b) tokenization may affect task performance: as LLMs are trained on tokenized text, they typically have some trouble identifying whether a letter is part of a word. For example, gpt-5-mini, after correctly identifying the rule about the letter p, remained unaware of the rule regarding the letter 'm', but remarked that there should be another rule, as "several rejected words do not contain 'p' (camera, tomato, cucumber, human, potato, pear, grape)" \textit{(sic)}. 

\section{More discussion on the ALP example}
\label{appendix_alp_discussion}
\renewcommand{\thefigure}{E\arabic{figure}}
\renewcommand{\thetable}{E\arabic{table}}
\setcounter{figure}{0}
\setcounter{table}{0}

Relevant notes on the ALP example are that the body of experimental data can easily be overwhelming, therefore our choice to rely on deterministic data-to-text generation and LLM-powered narrative generation, with the option for the agent to inspect the raw data (Figure \ref{fig1}b). However, with the longer experiments (7200 seconds) several issues arose, likely because of the length of the conversation. We included only experiments that were completed in this work.

Critical remarks for the ALP example include that even though LLM agents appear to use the right terminology, the exact meaning may be wrong to an expert. Hence it is important to check the actual experiments (specifically in experiment IIb). If available, the UMAP projection of the probed space against expert experiments (configuration III) provides us with additional validation that the agent did actually visit relevant parts of the experimental space. Furthermore, even though we made sure to strip the prompts of specific ALP terminology, providing the agent with only a sliver of additional information may (further) bias its results: we suspect that the term `monolayer' in configuration IIb steered the agent toward describing the process as `ALD'. 

Examples of other precautions we took were the removal of `precursor', `saturation', and a generic example recipe with only one pulse for the single-shot learning, in order to present the reactor as an open-ended tool. We did mention it was better to close a valve after opening it, but did not explicitly forbid leaving it open. Since LLM agents did explore co-dosing/decomposition, in one case ALD was not discovered, and cyclic pulsing is a logical step after discovering self-limiting surface-specific reactions, we believe LLM agents were not steered explicitly towards ALD-type reactions.

\end{document}